\documentclass[letterpaper, 10 pt, conference]{ieeeconf}  

\IEEEoverridecommandlockouts                              
\usepackage{hyperref}
\usepackage{url}
\usepackage{cite}
\usepackage{amsmath}
\usepackage{siunitx}
\usepackage{graphicx}
\usepackage{algorithm}
\usepackage{algpseudocode}
\usepackage{balance}
\newcommand{\HRule}{\rule{\linewidth}{0.25mm}}

\usepackage{color}

\newcommand{\update}[1]{\textcolor{black}{#1}}

\title{\LARGE \bf
Safe and Robust Motion Planning for Dynamic Robotics via Control Barrier Functions
}

\author{Aniketh Manjunath and Quan Nguyen
\thanks{This work is supported by USC Viterbi School of Engineering startup funds.}
\thanks{A. Manjunath is with the Department of Computer Science, University of Southern California, Los Angeles, California, 90089,
        {\tt\small v.m.aniketh@gmail.com.}}%
\thanks{Q. Nguyen is with the Department of Aerospace and Mechanical Engineering,
        University of Southern California,
        Los Angeles, California, 90089,
        {\tt\small quann@usc.edu.}}%
}

\begin{document}

\maketitle
\thispagestyle{empty}
\pagestyle{empty}

\begin{abstract}
Control Barrier Functions (CBF) are widely used to enforce the safety-critical constraints on nonlinear systems. Recently, these functions are being incorporated into a path planning framework to design safety-critical path planners. However, these methods fall short of providing a realistic path considering both the algorithm's run-time complexity and enforcement of the safety-critical constraints. This paper proposes a novel motion planning approach using the well-known Rapidly Exploring Random Trees (RRT) algorithm that enforces both CBF and the robot Kinodynamic constraints to generate a safety-critical path. The proposed algorithm also outputs the corresponding control signals that resulted in the obstacle-free path. The approach also allows considering model uncertainties by incorporating the robust CBF constraints into the proposed framework. Thus, the resulting path is free of any obstacles and accounts for the model uncertainty from robot dynamics and perception. Result analysis indicates that the proposed method outperforms various conventional RRT-based path planners, guaranteeing a safety-critical path with minimal computational overhead. We present numerical validation of the algorithm on the Hamster V7 robot car, a micro autonomous Unmanned Ground Vehicle that performs dynamic navigation on an obstacle-ridden path with various uncertainties in perception noises and robot dynamics.
\end{abstract}

\section{INTRODUCTION}
Path planning is a fundamental problem in robotics that involves moving a robot from an initial configuration to a goal configuration while avoiding obstacles along the way. For efficient navigation of autonomous mobile robots, we need an effective path planning algorithm that runs in real-time. As the search space is continuous, it is challenging to use the shortest path algorithms like Dijkstra's \cite{dijkstra}, or A* \cite{A_star} from graph theory.  One of the commonly used approaches is the Rapidly Exploring Random Trees (RRT) algorithm, which iteratively builds the path by randomly sampling the search space \cite{justRRT}. Although RRTs do not guarantee an optimal solution, it is relatively faster than other path planning algorithms. However, some robotic systems use a heuristic \cite{D_star, D_star_focused, D_star_lite, ARA_star} or potential field \cite{potential, potential_lim, harmonic_potential} approach to path planning. For autonomous systems with dynamic constraints, heuristic-based methods fall short on response time and cannot look ahead, resulting in instability and non-optimal paths. Agent-based RTT such as RRT* \cite{RRT_star} and RRT*-smart \cite{RRT*_smart} improves the quality of the path, but this comes at the cost of computation efficiency. Constrained RRT algorithms find various applications such as robotic manipulation \cite{CBiRRT}, or quadcopter navigation \cite{LazyRRT_star}, etc. Research on complex methods such as Probabilistic Roadmaps \cite{prm}, Machine Learning-based path planning approach \cite{mlpath}, Particle Swarm Optimization \cite{swarmplan} are gaining high interests, but present high computational complexities. 

Robot path planning has been under constant development, and it is challenging to devise algorithms that can ensure stability while satisfying safety-critical constraints for nonlinear systems. Recently, Control Barrier Functions (CBF) \cite{CBFs} was introduced to effectively address these constraints for nonlinear systems.  The approach has been successfully validated in different application including autonomous vehicle \cite{adaptivecruise, son2019safety}, or bipedal locomotion \cite{bipedalrobot, nguyen2020dynamic}. However, this is a feedback control framework and it does not consider long-term planning under the presence of multiple obstacles. Recently, a QP-formulation was introduced to enforce CBF constraints on the RRT algorithm to achieve obstacle avoidance \cite{CBF_RRT}. However, this work simulates the closed-loop system using CBF control inside the planning framework, therefore computationally expensive.

In this work, we proposed an effective framework to incorporate CBF constraints into the RRT motion planning approach to address the problem of dynamic robotics navigating in cluttered environments. Our work only considers CBF as a constraint of the RRT algorithms and therefore is balanced between safety guarantee and computation efficiency.
In addition, this framework also allows us to incorporate our prior work on robust CBF \cite{robustcbfquan} into this proposed approach to address the problem of motion planning under model uncertainty.

Following are the main contributions of this paper:
\begin{itemize}
    \item Introduction of a new technique on efficient path planning based on RRT and Kinodynamic Barrier Function (KBF) for improving the efficiency and quality of the path planning framework. 
    \item Introduction of a new technique on robust motion planning based on RRT and robust KBF for a flexible and systematic approach to address model uncertainties.
    \item Combination of robust motion planning and real-time safety-critical control to improve both high-level planning and real-time collision avoidance for dynamic robotics.
    \item Application to the problem of robust path planning and control for collision avoidance of autonomous vehicles under different levels of uncertainties.
\end{itemize}

The rest of the paper is organized as follows.
In Section \ref{S2}, we revisit the Quadratic Program (QP) formulation of Control Lyapunov Functions and Control Barrier Functions. This section also provides an overview of the Rapidly Exploring Random Trees (RRT) algorithm. We then present in Section \ref{sec:proposed_work} the proposed safety-critical path planner, followed by the integration of robustness into the same planning framework. Section \ref{sec:result} presents the numerical validation of the proposed work with the application to autonomous vehicles. Here, we also discuss the key advantages of the proposed work. Finally, we end with the concluding remarks.

\section{BACKGROUND}
\label{S2}
In this section, we present the details of car dynamics, QP formulation of different controllers followed by the RRT algorithm, all necessary features required for establishing the proposed work.

\subsection{Car Model}
We consider the bicycle model for car-like vehicles. We model the state of the robot as: 
\begin{eqnarray}
    &z = \begin{bmatrix} x_p & y_p & \theta & v \end{bmatrix} ^ T \\
    &\dot{z} = \begin{bmatrix} v\;cos\theta & v\;sin\theta & \; \update{v \frac{tan(\psi)}{L}}  & a \end{bmatrix} ^ T
\end{eqnarray}

Here $x_p, y_p, \theta, v$ corresponds to the robot's position, heading, and velocity, respectively, while $\psi$, $L$, and $a$ represent the steering angle, robot length, and acceleration, respectively. 
We define control $u$ with $c = \tan(\psi) / L$ as: 
\begin{equation}
    u = [c, a]^T.
    \label{eq:control}
\end{equation}

Using a simple transformation we define $x = \begin{bmatrix} x_1 & x_2 \end{bmatrix} ^ T$ with 
$x_1 = \begin{bmatrix} x_p & y_p \end{bmatrix} ^ T$ and 
$x_2 = \begin{bmatrix} v_x & v_y \end{bmatrix} ^ T$. 

By computing the derivatives $\Dot{x}_2 = \begin{bmatrix}\Dot{v}_x &  \Dot{v}_y\end{bmatrix}^T$, we get:

\begin{eqnarray}
\Dot{x}_2 = 
        \begin{bmatrix}
            -v^2\sin{\theta} & \cos{\theta} \\
            v^2\cos{\theta} & \sin{\theta}
        \end{bmatrix}
        \times
        \begin{bmatrix}
            \dfrac{\tan{\psi}}{L} \\ 
            a
        \end{bmatrix}.
\end{eqnarray}

The system dynamics now become:
\begin{equation}
    \dot{x}_1 = x_2 \;;\; \dot{x}_2 = f(x) + g(x) u, \label{eq:robot_dynamics}
\end{equation}
where
\begin{equation} 
    f(x) = \begin{bmatrix}
                0 \\
                0
            \end{bmatrix}
    \text{ and }
    g(x) = \begin{bmatrix}
            -v^2 \sin{\theta} & \cos{\theta} \\
            v^2 \cos{\theta} & \sin{\theta}
           \end{bmatrix}.
\end{equation}

We now can apply input-ouput linearization \cite{RES_CLF} using the following pre-control law:
\begin{equation}
    u = g(x)^{-1}(\mu - f(x)).
    \label{eq:pre-control}
\end{equation}

Given the current state $x$ and a reference state $x_{rm}$, we define error as:
\begin{equation} 
    e = x_{rm} - x \label{eq:err}.
\end{equation}

Deriving the I-O linearized dynamics, we get:
\begin{equation} 
    \dot{e} = Fe + G\mu,
\end{equation}
\indent where
\begin{equation}
    \text{F = } 
            \begin{bmatrix}
                0 & I \\
                0 & 0 \\
            \end{bmatrix}
        \text{ and G = }
        \begin{bmatrix}
            0 \\
            I \\
        \end{bmatrix}.
\end{equation}

The system can be stabilized using a simple PD control:
\begin{equation}
    \mu_{pd} = \begin{bmatrix}-K_P & -K_D\end{bmatrix}e,
\end{equation}
with $K_P$ and $K_D$ as proportional and derivative constants. 
\subsection{Control Lyapunov Function (CLF)}
Having presented the dynamics for autonomous vehicles, we now present a feedback control framework based con CLF-CBF-QP to guarantee stability and safety-critical constraints for the control system.

We first begin by generating a Control Lyapunov Function (CLF) \cite{RES_CLF}, by solving for $P$ in the following Lyapunov equation:
\begin{equation}
    A^T P + PA = -Q,    
\end{equation}

where
\begin{equation}
    Q = I \text{ ;  } A = \begin{bmatrix}0 & I \\ -K_P & -K_D \end{bmatrix}.
\end{equation}

We now define the CLF for our system (\ref{eq:robot_dynamics}) as:
\begin{eqnarray}
        &V = e^T Pe \\ 
        &L_f V = e^T (F^T P + PF) e \\ 
        &L_g V = 2 e^T PG
\end{eqnarray}
where $L_f$ and $L_g$ are the Lie derivatives of functions $f$ and $g$, respectively. This gives us the CLF condition that enforces stability on the system:
\begin{equation}
    L_f V + L_g V\mu + e^T Q e \leq 0.
    \label{eq:clf}
\end{equation}

By enforcing equation (\ref{eq:clf}) in our Quadratic Program, we have our CLF-QP controller:

\noindent\HRule\\
\noindent \textbf{CLF-QP:}
\begin{eqnarray*}
    &\min ||\mu - \mu_{pd}||^2 \\
    &\text{s.t.} -\infty \leq L_g V\mu \leq -L_f V - e^T Q e \text{ (\bf CLF\bf)}
\end{eqnarray*}
\HRule

\subsection{Exponential Control Barrier Function (CBF)}
The CLF-QP controller discussed earlier only assures stability and does not guarantee safety-critical constraints, such as collision avoidance. Given the current position of the robot $({x_p}, {y_p})$ and obstacle of known radius $r_o$ at position $({x_o}, {y_o})$, we can define the Barrier function $B(x)$ that strictly enforces safety-critical constraint \cite{ECBF}. For simplicity, we set a safety boundary for the car using a circle with the radius $r_r$. Hence, the safety distance between the robot and the obstacle is set to $r = r_o + r_r$. In this paper, we will use this distance to maintain collision-free between the car and surrounding obstacles. We then can enforce safety using Exponential CBF \cite{ECBF} as follows:

In order to guarantte:
\begin{eqnarray}
    B(x) = ({x_p}-{x_o})^2 + ({y_p} - {y_o})^2 -r^2 \ge 0, 
\end{eqnarray}
we can enforce the following constraint:
\begin{equation}
    {B_1}(x) = {\dot{B}(x)} + {\gamma_1}{B(x)} \ge 0\;\;\;(with\;\;{\gamma_1} > 0),
\end{equation}
where
\begin{eqnarray}
    \Dot{B}(x) = 2(x_p - x_o)v_x + 2(y_p - y_o)v_y.
\end{eqnarray}

By repeating the step one more time, we arrive at the following CBF condition:
\begin{equation}
    \dot{B_1}(x) + {\gamma_2}{B_1(x)} \ge 0\;\;\;\;(with\;\;{\gamma_2} > 0),
    \label{eq:cbf}
\end{equation}
where
\begin{equation}
    \Dot{B}_1(x) = \gamma_1\Dot{B}(x) + 2v_x^2 + 2v_y^2 
                    + 2 \begin{bmatrix}x_p - x_o & y_p - y_o\end{bmatrix} \mu.
\end{equation}

Adding (\ref{eq:cbf}) into the set of QP constraints, we derive our CLF-CBF-QP Controller:

\noindent\HRule\\
\noindent \textbf{CLF-CBF-QP:}
\begin{eqnarray*}
    &\min ||\mu - \mu_{pd}||^2 + Pd^2 \\
    &\text{s.t.} -\infty \leq L_f V + L_g V\mu + e^T Q e \leq d \text{ (\bf CLF\bf)} \\
    &0 \leq \Dot{B_1}(x) + \gamma_2B_1(x) \leq \infty \text{ (\bf CBF\bf)}
\end{eqnarray*}
\HRule

Here, the additional term $Pd^2$ is introduced to relax the bound of the CLF condition for satisfying conflicting additional constraints. Here, $d > 0$ and $P$ is a large positive number that represents the penalty for relaxing the inequality constraints.
\subsection{Rapidly-exploring Random Tree (RRT)}
In the previous sections, we discussed various controllers that minimize error while enforcing stability and safety-critical conditions. To compute error $e$, we need a reference model $x_{rm}$, see equation (\ref{eq:err}). Given the start and goal state, we need a path planner that returns the required reference $x_{rm}$. Although RRTs does not guarantee optimality, they are comparatively faster than other methods. Hence, in this work, we use RRTs for path planning, see Algorithm~\ref{alg:rrt}.

Within the defined search space, the RRT algorithm grows a tree from start to goal by performing the following three functions where each node is associated with the state as given by the system dynamics: 

\begin{itemize}
\item {\bf{get\_random\_sample}} returns a random sample in the search space with a uniform probability distribution. 
\item {\bf get\_nearest\_neighbor\bf} returns the nearest neighbor of the random sample from the set of previously visited samples of the tree based on the Euclidean distance.  
\item {\bf extend\_sample\bf} checks if the path from the nearest neighbor node and the random sample is free of obstacles. The Auxillary function {\bf check\_collision\bf} is used to determine whether the new state violates the safety constraints. If the constraint is satisfied, the trees grow by constant step-size in the direction of the random sample. In presence of more than one obstacle, this step is repeated for each obstacle and the new state is considered valid if it satisfies the safety constraints for all obstacles.
\end{itemize}

The RRT tree iteratively grows until it reaches the goal state. As an added measure, to limit the time, the total number of iterations is limited. \update{In the next section, we will use the knowledge of CLF and CBF controllers to formulate our safety-critical path planner.}

\begin{algorithm}[t]
\caption{RRT}\label{alg:rrt}
\begin{algorithmic}
\Function{RRT}{$start_{state}, goal_{state}, step\_size, obstacles$}
    \State $tree \gets (null, start_{state})$
    \State $s \gets step\_size$
    \State $o \gets obstacles$
    \While{$tree.visited(goal_{state})$ \text{is} $False$}       
        \State {$node_{r} \gets \bf get\_random\_sample()\bf$}
        \State {$node_{n}$ $\gets \bf extend\_sample\bf${($tree, node_{r} , s, o$)}}
        \State {if $node_{n}$ \text{ it not } $null$}
        \State \hspace{5mm} $tree.add(node_{n}, node_{r})$
    \EndWhile
    \State \textbf{return} $tree$
\EndFunction

\\

\Function{extend\_sample}{$tree,node_r, s, o$}
    \State {$node_n$ $\gets \bf get\_nearest\_neighbor\bf${($tree, node_r$)}}
    \State {if $\bf{check\_collision}$ $(node_n, node_r, o)$}
    \State \hspace{5mm} \textbf{return} $null$\Comment{path has obstacles}
    \State {else}
    \State \hspace{5mm} \textbf{return} $node_n$\Comment{path is obstacle free}
\EndFunction
\end{algorithmic}
\end{algorithm}

\section{PROPOSED WORK}
\label{sec:proposed_work}
Having presented the background on RRTs and various controllers, this section presents our proposed work on designing a path planner that incorporates safety-critical constraints into the path planning framework. We introduce a safety-critical path planner that runs in real-time, even under the influence of uncertainties. We first present RRT-KBF aimed at improving the efficiency and quality of the path planning framework, followed by Robust RRT-KBF that adds robustness to the previous method.

\subsection{RRT with Kinodynamic Barrier Function (RRT-KBF)}

Tracing back to the RRT algorithm, the first step involves fetching a random sample. Given the current state and the random sample, we check if the path connecting them is safe. A QP-solver can be used to find an optimal control $\mu$ that enforces CBF constraints \cite{CBF_RRT}. However, using the QP-solver for every iteration significantly delays the convergence of RRT. Therefore, we propose an alternative method for finding the safety-critical control,  see Algorithm~\ref{alg:rrt-kbf}. We first select a random node from the set of visited tree nodes. Owning to the random nature of RRT, we also generate a random control $u$ from a uniform distribution of kinodynamically constrained control. The required control (\ref{eq:control}) is obtained from:
\begin{equation}
    u = \begin{bmatrix} \frac{\tan\psi}{L} \\ a \end{bmatrix} = \begin{bmatrix}
        uniform\;(\frac{-\tan\psi_{max}}{L}, \frac{\tan\psi_{max}}{L}) \\ uniform\;(0, a_{max})
    \end{bmatrix},
\end{equation}
where $L$, $\psi_{max}$, and $a_{max}$ are the length, maximum steering angle, and acceleration of the robot. Given a random state and the random control $u$, we determine if the control satisfies the CBF constraints (\ref{eq:cbf}). If the control $u$ satisfies the safety-critical condition, we apply the control $u$ for a fixed time step $dt$ to reach the new state. This process repeats until we reach the goal state. By preserving the random nature of RRT, the proposed algorithm can achieve a faster path that satisfies the safety-critical constraint.

By introducing the CBF constraints (\ref{eq:cbf}) in the exploration phase of RRT, the proposed work ensures that every valid state of the tree enforces the safety criterion. Hence, the overall path traced from the start state to the goal state is safe.

\begin{algorithm}[t]
\caption{RRT-KBF}\label{alg:rrt-kbf}
\begin{algorithmic}
\State $F, G, \psi_{max}, a_{max}, L, \gamma_1, \gamma_2 \gets constants$
\Function{RRT}{$start_{state}, goal_{state}, obst$}
    \State $tree \gets (null, start_{state}, null)$
    \While{$tree.visited(goal_{state})$ \text{is} $False$}       
        \State {$node_{v} \gets \bf get\_random\_visited\_sample\bf \text{($tree$)}$}
        \State {$node_{n}, u$ $\gets \bf extend\_sample\bf${($node_{v}, obst$)}}
        \State {if $node_{n}, u$ \text{ it not } $null$}
        \State \hspace{5mm} $tree.add(node_{n}, node_{r}, u)$
    \EndWhile
    \State \textbf{return} $tree$
\EndFunction

\\

\Function{extend\_sample}{$node_n, obst$}
    \State {$u \gets \bf get\_uniform\_sample()\bf$}
    \State {if $\bf{check\_collision\_kbf\bf}(node_n, u, obst)$}
    \State \hspace{5mm} \textbf{return} $null, null$\Comment{$u$ is non-optimal control}
    \State {else}
    \State \hspace{5mm} \textbf{return} $node_r, u$\Comment{$u$ is optimal control}
\EndFunction

\\

\Function{check\_collision\_kbf}{$node, u, obst$}
    \State $x_p, y_p \gets node.state$
    \State $x_o, y_o, r_o \gets obst.state$
    \State $v_x, v_y \gets node.vel.x, node.vel.y$
    \State $B \gets ({x_p}-{x_o})^2 + ({y_p} - {y_o})^2 -r_o^2$
    \State $\dot{B} \gets 2(x_p-x_o)v_x + 2(y_p - y_o)v_y$
    \State ${B_1} \gets {\dot{B}} + {\gamma_1}{B}$
    \State $\mu \gets f(x) + g(x)u$
    \State $\Dot{B_1} \gets \gamma_1\Dot{B} + 2v_x^2 + 2v_y^2 
                    + 2 \begin{bmatrix}x_p - x_o & y_p - y_o\end{bmatrix} \mu$
    \State if $\Dot{B_1} + \gamma_2B_1 < 0$
    \State \hspace{5mm} \textbf{return} $True$ \Comment{CBF constraint fail}
    \State {else}
    \State \hspace{5mm} \textbf{return} $False$ \Comment{CBF constraint pass}
\EndFunction

\end{algorithmic}
\end{algorithm}

\subsection{Robust RRT with Kinodynamic Barrier Function (Robust RRT-KBF)}

\begin{algorithm}[!t]
\caption{Robust RRT-KBF}\label{alg:r-rrt-kbf}
\begin{algorithmic}
\State $F, G, \psi_{max}, a_{max}, L, \gamma_1, \gamma_2, \Delta_1, \Delta_2 \gets constants$
\Function{check\_collision\_robust\_kbf}{$node, u, obst$}
    \State $x_p, y_p \gets node.state.pos$
    \State $x_o, y_o \gets obst.state$
    \State $v_x, v_y \gets node.state.vel.x, node.state.vel.y$
    \State $B \gets ({x_p}-{x_o})^2 + ({y_p} - {y_o})^2 -r^2$
    \State $\dot{B} \gets 2(x_p-x_o)v_x + 2(y_p - y_o)v_y$
    \State ${B_1} \gets {\dot{B}} + {\gamma_1}{B}$
    \State $\mu \gets  F + Gu$
    \State $\Dot{B_1} \gets \gamma_1\Dot{B} + 2v_x^2 + 2v_y^2 
                    + 2 \begin{bmatrix}x_p - x_o & y_p - y_o\end{bmatrix} \mu$
    \State $A \gets \gamma_1\dot{B_1} + 2v_x^2 + 2v_y^2 + \gamma_2B_1$
    \State $b \gets 2 [x_p - x_o \;\; y_p - y_o]$
    \State $\psi_{0} \gets max{((A + b * \Delta_{1}\textsuperscript{max}),(A - b * \Delta_{1}\textsuperscript{max}))}$ 
    \State $\psi_{p} \gets b * (1 + \Delta_{2}\textsuperscript{max})$
    \State $\psi_{n} \gets b * (1 - \Delta_{2}\textsuperscript{max})$
    \State if $\psi_{0} + \psi_{p} * \mu \le 0 \text{ and } \psi_{0} - \psi_{n} * \mu \le 0$
    \State \hspace{5mm} \textbf{return} $False$ \Comment{Robust constraint pass}
    \State else
    \State \hspace{5mm} \textbf{return} $True$ \Comment{Robust constraint fail}
\EndFunction
\end{algorithmic}
\end{algorithm}

The CBF equation (\ref{eq:cbf}) discussed in the previous section assumes that the true dynamics of the system are known. To handle model uncertainties, QP formulations of CBF were extended to yield the robust CBF constraints, as introduced in \cite{robustcbfquan}. We extend this work, deriving the robust CBF constraint for the previously proposed RRT-KBF path planner, see Algorithm~\ref{alg:r-rrt-kbf}. To account for the model uncertainties, we can use nominal dynamics model $\Tilde{f}(x)$ and $\Tilde{g}(x)$ in place of unknown true dynamics $f(x)$ and $g(x)$. Now, equation (\ref{eq:pre-control}) can be rewritten as:
\begin{equation}
    \Tilde{u} = \Tilde{g}(x)^{-1}(\mu - \Tilde{f}(x))
    \label{eq:robust-pre-control}
\end{equation}
Substituting (\ref{eq:robust-pre-control}) into the dynamics equation (\ref{eq:robot_dynamics}) we get:
\begin{equation}
    \dot{x_2} = \mu + \Delta_1 + \Delta_2\mu 
\end{equation}
\indent where
\begin{align}
    &\Delta_1 = f(x) - g(x) \Tilde{g}(x)^{-1}\Tilde{f}(x) \\
    &\Delta_2 = g(x)\Tilde{g}(x)^{-1} - \mathrm{I}
\end{align}
Here $\Delta_1$ and $\Delta_2$ represents the model uncertainty. Given pseudo-control $\mu$, the CBF equation (\ref{eq:cbf}) can be expressed as:
\begin{equation}
    \dot{B_1}(x, \mu) = A(x) + b(x)\mu \ge 0 \label{eq:cbf_mu}
\end{equation}
\indent where
\begin{eqnarray}
    &A(x) = \gamma_1\dot{B}(x) + 2v_x^2 + 2v_y^2 + \gamma_2B_1(x)\\
    &b(x) = 2\ [x_p - x_o \;\; y_p - y_o]
\end{eqnarray}

Now equation (\ref{eq:cbf_mu}) becomes:
\begin{equation}
    \dot{B_1}(x, \mu) = A(x) + b(x) (\mu + \Delta_1 + \Delta_2\mu) \ge 0
    \label{eq:cbf_uncertain}
\end{equation}

To minimize model uncertainties, we have to minimize equation (\ref{eq:cbf_uncertain}). Considering positive constants $\Delta_{1}^{max}$, $\Delta_{2}^{max}$ with absolute values $|\Delta_1| \le \Delta_{1}^{max}$ and $|\Delta_2| \le \Delta_{2}^{max}$, we can rewrite (\ref{eq:cbf_uncertain}) as a maximization problem:
\begin{eqnarray}
    &\dot{B_1}(x, \mu) = A(x) + b(x) (\mu + \Delta_{1}^{max} + \Delta_{2}^{max}\mu)
    \label{eq:cbf_uncertain_max} \\
    &\dot{B_1}(x, \mu) = \max(\psi_0 + \psi_1\mu)  \le 0
\end{eqnarray}

The maximum values for this function is given by:
\begin{eqnarray}
    &\psi_0^{max} = \max
    \begin{cases}
        A(x) + b(x) \Delta_{1}^{max} \\
        A(x) - b(x)\Delta_{1}^{max}
    \end{cases} \\
    &\psi_1^{p} = b(x) (1 + \Delta_{2}^{max}) \\
    &\psi_1^{n} = b(x) (1 - \Delta_{2}^{max})
\end{eqnarray}

Finally, we have the robust CBF constraints that has accounted for model uncertainties:
\begin{eqnarray}
    &\psi_0^{max} + \psi_1^{p} \mu \le 0 \label{eq:rcbf_1}\\
    &\psi_0^{max} + \psi_1^{n} \mu \le 0 \label{eq:rcbf_2}
\end{eqnarray}

To incorporate the robust CBF controller into our framework, we modify the $\bf{check\_collision\_kbf\bf}$ in RRT-KBF algorithm to introduce robust constraints (\ref{eq:rcbf_1}) and (\ref{eq:rcbf_2}) in the path planner.

\section{NUMERICAL VALIDATION}
\label{sec:result}
\begin{figure}[!btp]
    \centering
    \includegraphics[width=0.48\textwidth]{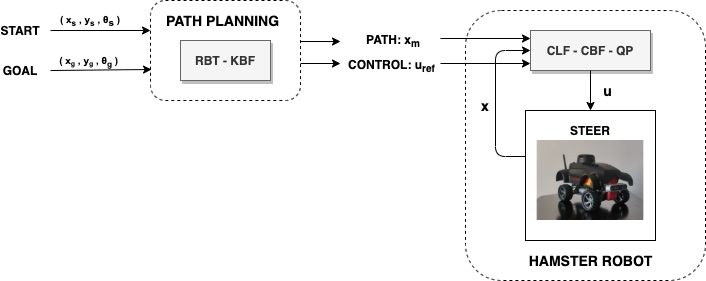}
    \caption{Block diagram of the proposed algorithm with application to autonomous vehicles.}
    \label{blk}
\end{figure}

To demonstrate the effectiveness of the proposed work, we design a two-stage pipeline involving the path-planning phase followed by a path-following phase. The path-planner (Algorithm \ref{alg:rrt-kbf}, \ref{alg:r-rrt-kbf}) generates a safety-critical reference path $x_{rm}$ with the corresponding control $u_{ref}$. This information is used by the path-follower with a feedback controller to generate a control that guarantees stability and safety-critical constraints for the autonomous vehicle. An overview of the simulation is presented as a block diagram in Fig.~\ref{blk}. 

We conducted the numerical simulation on Cogniteam's Hamster Robot \cite{HamsterV7}. The Hamster V7 robot uses 2 Raspberry Pi modules (master-slave setup) and Arduino Board as a low-level controller. This robot can move with a maximum velocity of \SI{1.2}{\meter/\second} and can make maximum turns of 30$^{\circ}$ angle. \update{We have abstracted obstacles as a circle for simplification. This may be extended for other geometric shapes.} The reader is encouraged to watch the supplemental video\footnote{\url{https://youtu.be/LtoM7CMWtp8}} 
for visualizations of our results.

\begin{figure}
    \centering
    $\begin{array}{cc}
        \includegraphics[width=0.21\textwidth]{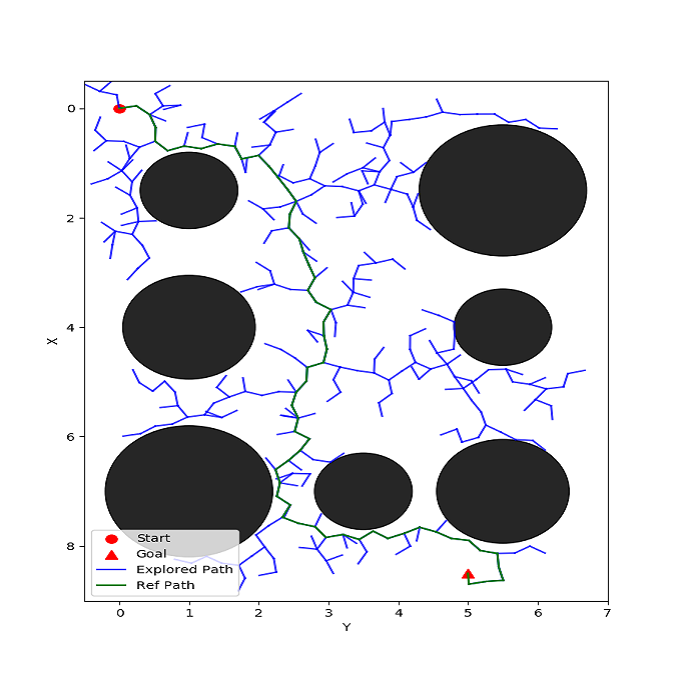}&
        \includegraphics[width=0.21\textwidth]{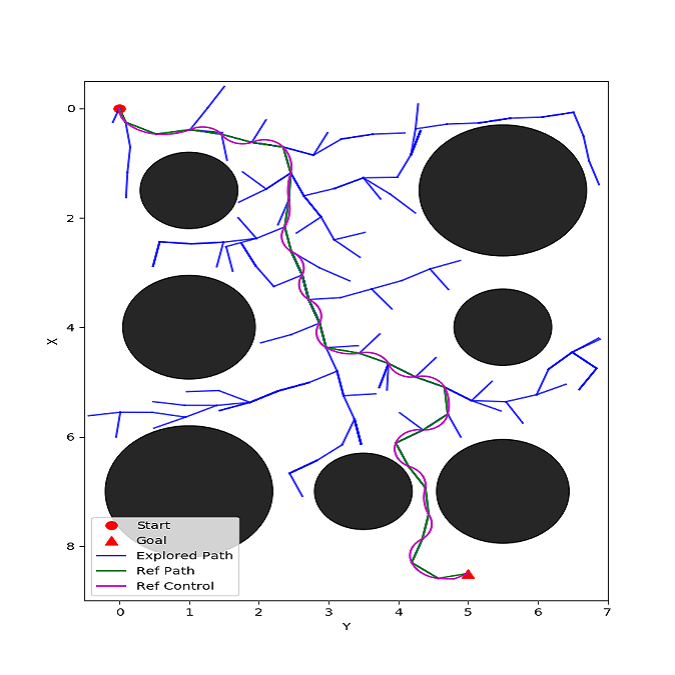}\\
        \mbox{(a1)}&\mbox{(b1)}\\
        \includegraphics[width=0.21\textwidth]{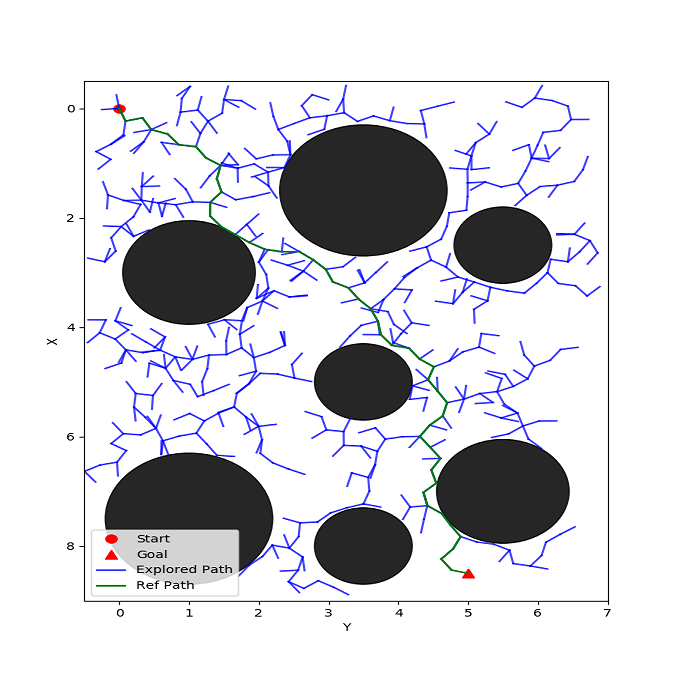}&
        \includegraphics[width=0.21\textwidth]{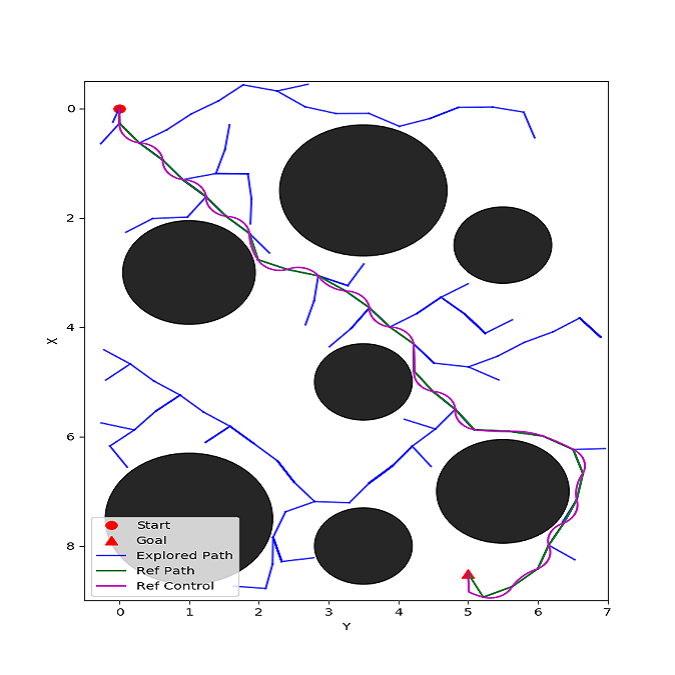}\\
        \mbox{(a2)}&\mbox{(b2)}\\
        \includegraphics[width=0.21\textwidth]{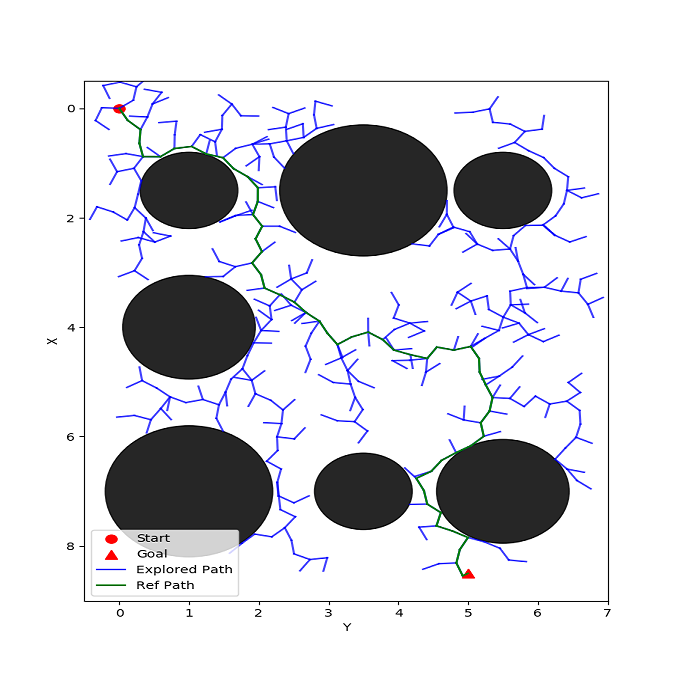}&
        \includegraphics[width=0.21\textwidth]{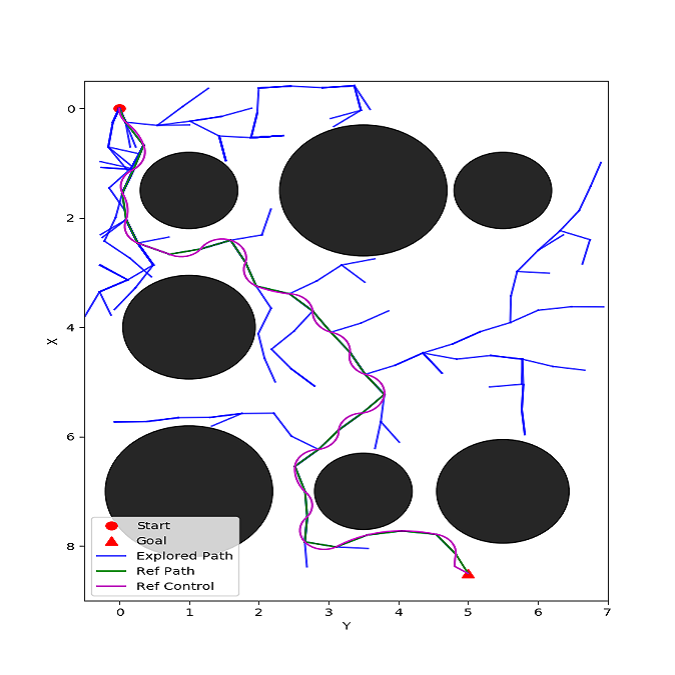}\\
        \mbox{(a3)}&\mbox{(b3)}\\
    \end{array}$
    \caption{Comparative analysis: [a1 - a3] showcase the performance of RRT, [b1 - b3]: showcase the performance of RRT-KBF on different scenarios, respectively}
    \label{comparison}
\end{figure}

\subsection{Path Planning}

In the first phase, we run the path planner by providing the start, goal, and obstacles information to Algorithm \ref{alg:rrt-kbf}. The path planner returns the reference model and the reference control as outputs. The path traced by different path planners is shown in Fig.~\ref{comparison}. There are several advantages of using the proposed RRT-KBF planner over RRT:

\begin{itemize}
    \item The path provided by RRT-KBF is kinematically constrained.
    \item The RRT-KBF algorithm also outputs the corresponding control that satisfies safety conditions (\ref{eq:cbf}). The control can be directly applied to the robot to take it from start to goal safely.
    \item The RRT algorithm uses fixed step size for path planning, whereas RRT-KBF uses dynamic steps based on the randomly sampled control. 
\end{itemize}

\subsection{Path Following}

\begin{figure}
    \centering
    \includegraphics[width=0.21\textwidth]{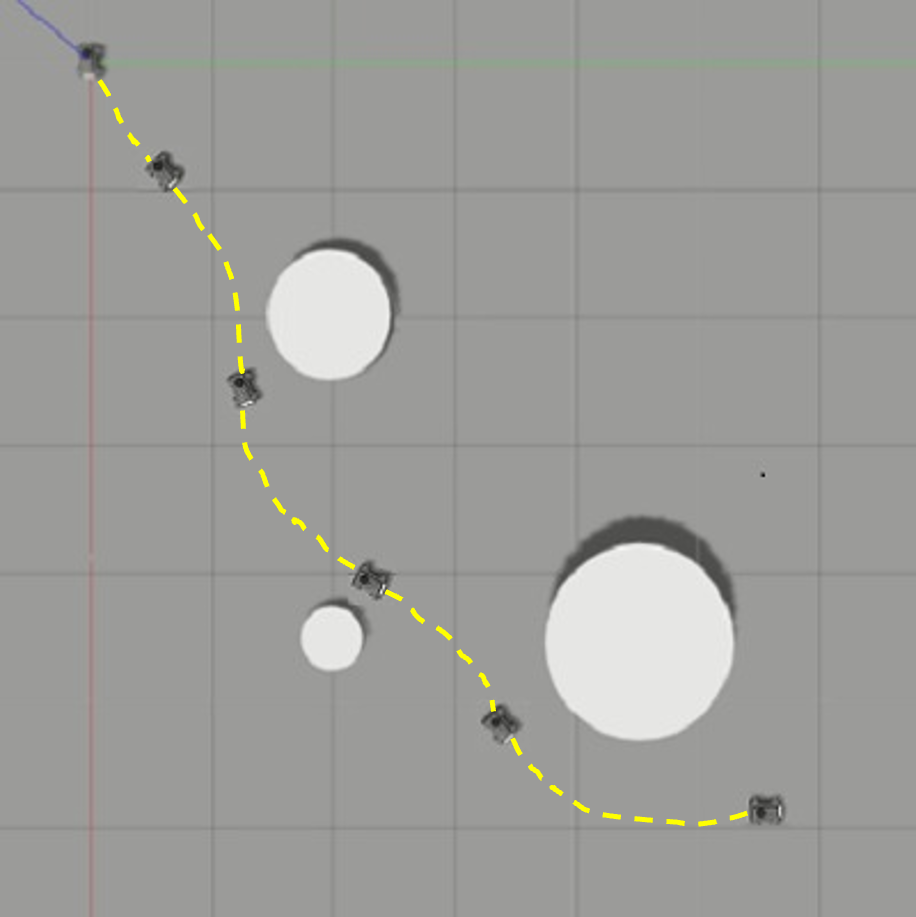}
    \caption{Snapshots of ROS simulation running RRT-KBF algorithm. Simulation video: \url{https://youtu.be/LtoM7CMWtp8}.}
    \label{fig:sim}
\end{figure}

The outputs of the proposed path planners enforce only safety-critical conditions (\ref{eq:cbf}). We need a feedback controller that satisfies the stability constraints (\ref{eq:clf}) running on the autonomous vehicle. The reference model and control returned by the path planners are used by the CLF-CBF-QP controller running on the Hamster robot to enforce stability constraints over the safety-critical path. This ensures that the final path traced by the Hamster robot is both stable and safe. 
Fig.~\ref{fig:sim} shows motion snapshots of the Hamster Robot using our proposed approach in ROS/Gazebo simulation.

\begin{table*}[!t]
\centering
\caption{Performance analysis of path planning algorithms}
\label{tab:comp}
\begin{tabular}{|l|c|c|c|c|c|c|c|}
\hline
                    & \multicolumn{3}{c|}{\textbf{CONSTRAINTS}}               & \multicolumn{4}{c|}{\textbf{AVG. TIME (s) OVER 100 RUNS}}                                                                                                                \\ \cline{2-8} 
                    & \textbf{SAFETY} & \textbf{DYNAMIC} & \textbf{KINEMATIC} & \multicolumn{1}{l|}{\textbf{SCENARIO 1}} & \multicolumn{1}{l|}{\textbf{SCENARIO 2}} & \multicolumn{1}{l|}{\textbf{SCENARIO 3}} & \multicolumn{1}{l|}{\textbf{SCENARIO 4}} \\ \hline
\textbf{RRT}        & YES             & NO               & NO                 & 0.71                                     & 0.77                                     & 0.73                                     & 0.77                                     \\ \hline
\textbf{RRT-CBF QP} & YES             & YES              & NO                 & 9.37                                     & 11.72                                    & 9.42                                     & 11.14                                    \\ \hline
\textbf{RRT-KBF}    & YES             & YES              & YES                & 1.11                                     & 1.12                                     & 1.16                                     & 1.18                                     \\ \hline
\end{tabular}
\end{table*}

\begin{table}[!t]
\centering
\caption{Performance analysis with model uncertainty}
\label{tab:rob_comp}
\begin{tabular}{l|c|c|c|}
\cline{2-4}
\multicolumn{1}{c|}{\textbf{}}                & \multicolumn{3}{c|}{\textbf{\begin{tabular}[c]{@{}c@{}}AVG. TIME (s) OVER 100 RUNS\end{tabular}}}                                                                                                                                                                       \\ \hline
\multicolumn{1}{|c|}{\textbf{Robust RRT-KBF}} & \multicolumn{1}{l|}{\textbf{\begin{tabular}[c]{@{}l@{}}$\Delta_{1}^{max}$ = 0\\ $\Delta_{2}^{max}$ = 0\end{tabular}}} & \multicolumn{1}{l|}{\textbf{\begin{tabular}[c]{@{}l@{}}$\Delta_{1}^{max}$ = 0.3\\ $\Delta_{2}^{max}$ = 30\end{tabular}}} & \multicolumn{1}{l|}{\textbf{\begin{tabular}[c]{@{}l@{}}$\Delta_{1}^{max}$ = 0.5\\ $\Delta_{2}^{max}$ = 50\end{tabular}}} \\ \hline
\multicolumn{1}{|l|}{\textbf{SCENARIO 1}}     & 1.11                                                                                  & 1.12                                                                                     & 3.06                                                                                     \\ \hline
\multicolumn{1}{|l|}{\textbf{SCENARIO 2}}     & 1.12                                                                                  & 1.22                                                                                     & 3.24                                                                                     \\ \hline
\multicolumn{1}{|l|}{\textbf{SCENARIO 3}}     & 1.16                                                                                  & 1.10                                                                                     & 3.38                                                                                     \\ \hline
\multicolumn{1}{|l|}{\textbf{SCENARIO 4}}     & 1.18                                                                                  & 1.24                                                                                     & 3.44                                                                                     \\ \hline
\end{tabular}
\end{table}

\subsection{Robustness to Uncertainties}

\begin{figure}[!h]
    \centering
    $\begin{array}{cc}
        \includegraphics[width=0.21\textwidth]{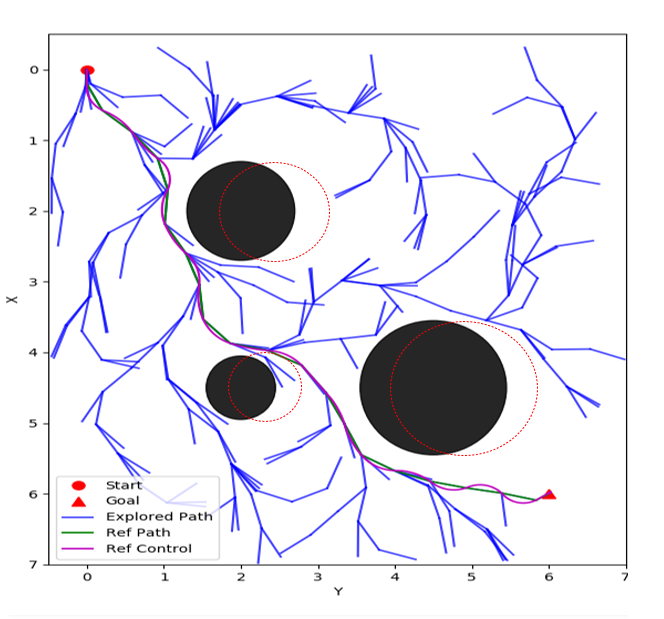}&
        \includegraphics[width=0.21\textwidth]{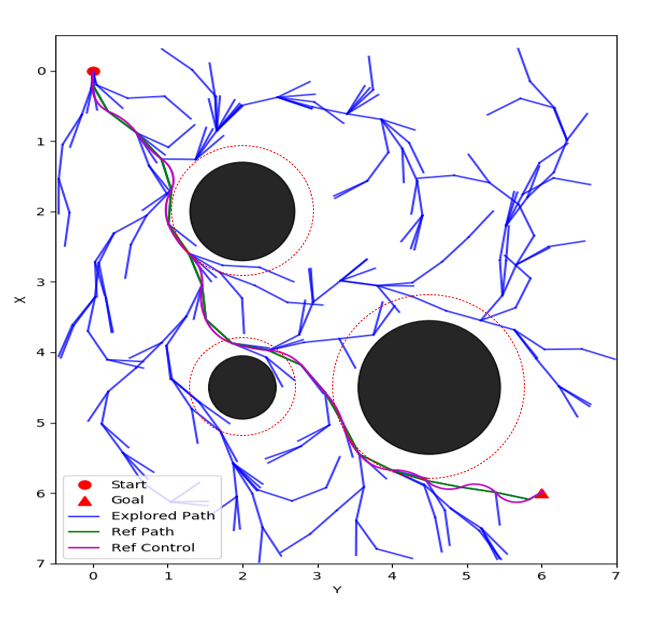}\\
        \mbox{(a1)}&\mbox{(a2)}\\
        \includegraphics[width=0.21\textwidth]{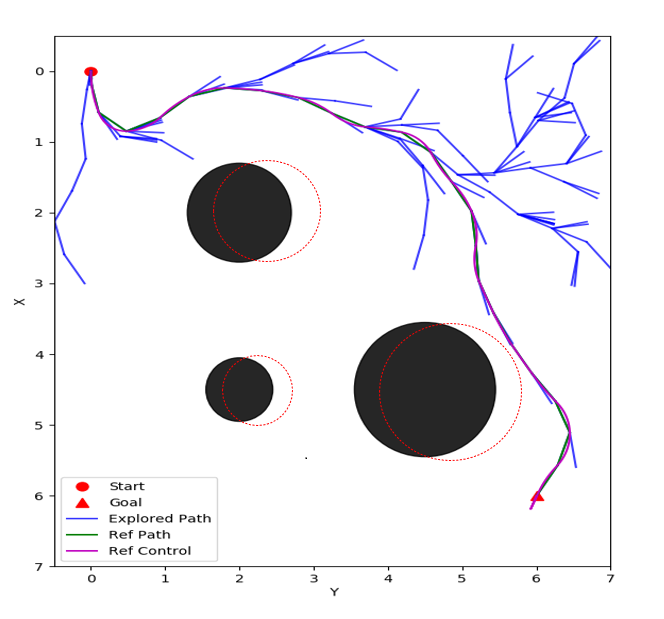}&
        \includegraphics[width=0.21\textwidth]{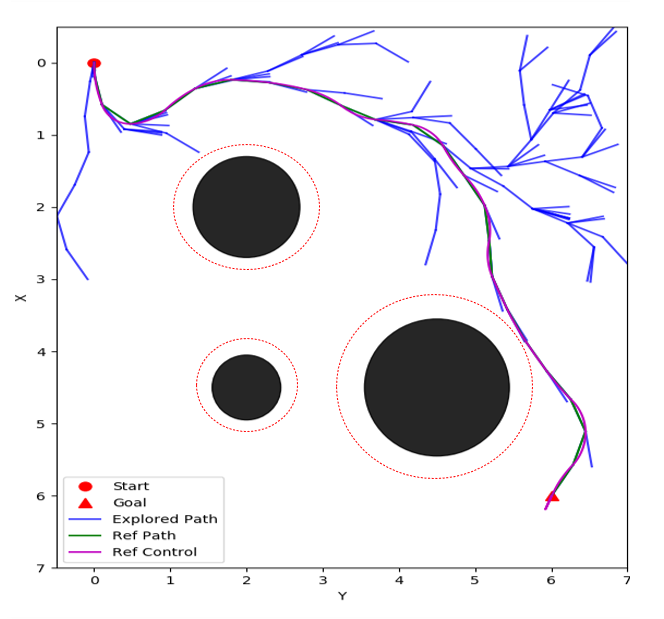}\\
        \mbox{(b1)}&\mbox{(b2)}\\
    \end{array}$
    \caption{Comparative analysis: [a1 - a2] showcase the performance of RRT-KBF, [b1 - b2] showcase the performance of Robust RRT-KBF with position and radius uncertainties of 0.5m and 0.25m, respectively.}
    \label{fig:cmp_unc}
\end{figure}

The Robust RRT-KBF uses the Robust CBF constraints (\ref{eq:rcbf_1}, \ref{eq:rcbf_2}) in the path planning framework to can handle model uncertainties. To evaluate the performance of the algorithm under uncertainty, we introduce an error of 0.5m in the position of the obstacles and an error of 0.25m in the size/radius of the obstacle.
Fig.~\ref{fig:cmp_unc} shows the effectiveness of Robust RRT-KBF \update{($\Delta_{1}^{max} = 10$, $\Delta_{2}^{max} = 0.1$)} over RRT-KBF in both the scenarios.  

\begin{figure}[!btp]
    \centering
    $\begin{array}{cc}
        \includegraphics[width=0.21\textwidth]{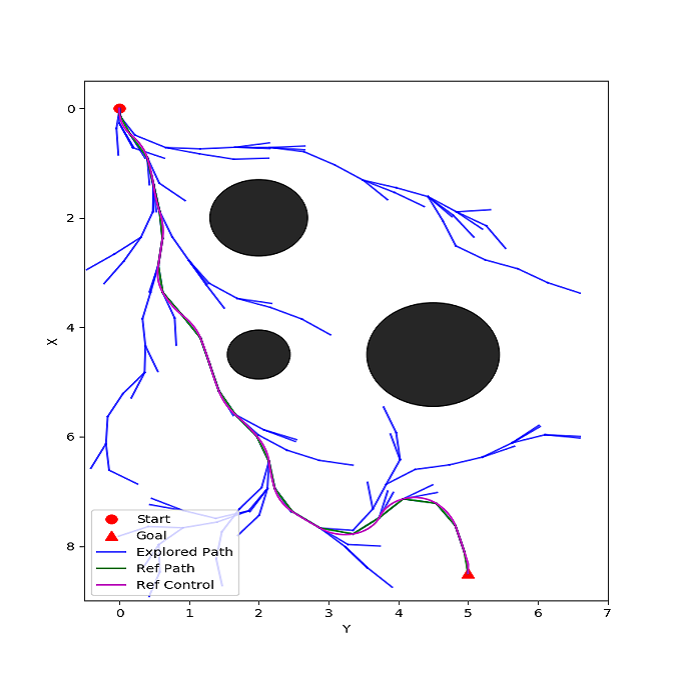}&
        \includegraphics[width=0.21\textwidth]{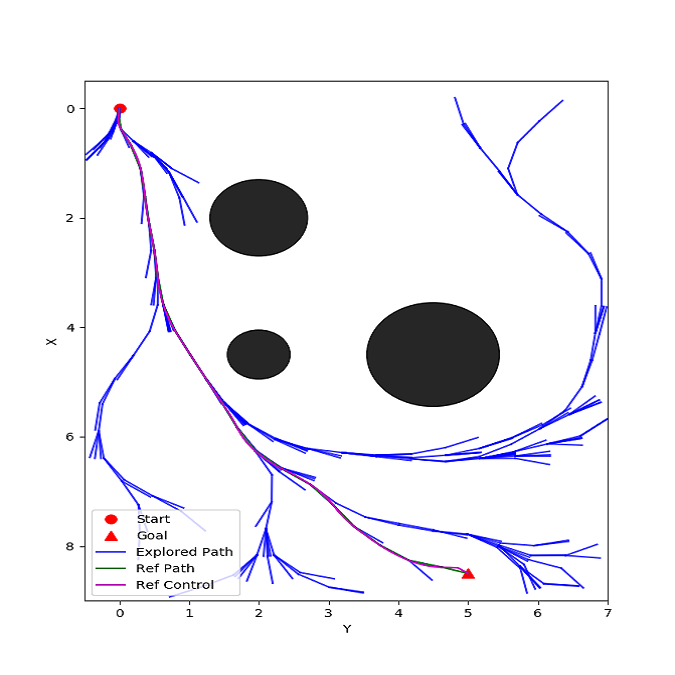}\\
        \mbox{(a) \small$\Delta_1^{max} = 0.1, \Delta_2^{max} = 10$}&\mbox{(b) \small$\Delta_1^{max} = 0.2, \Delta_2^{max} = 20$}\\
        \includegraphics[width=0.21\textwidth]{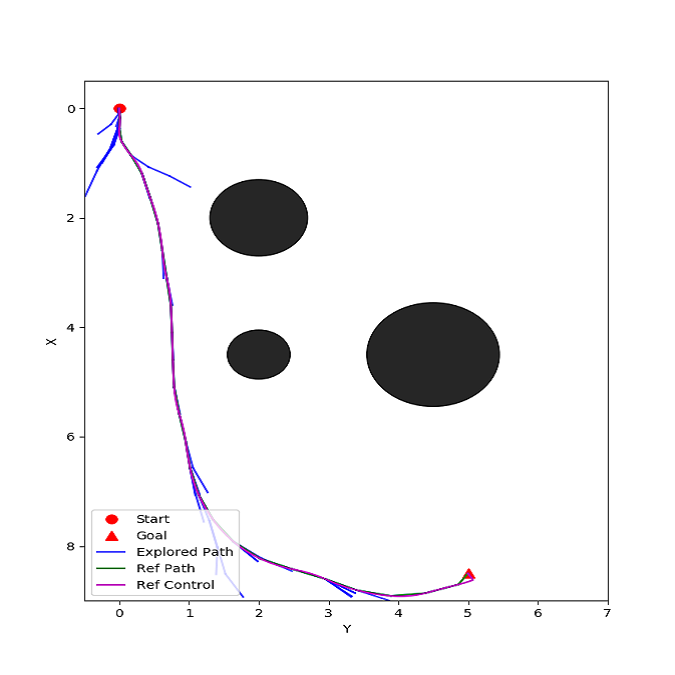}&
        \includegraphics[width=0.21\textwidth]{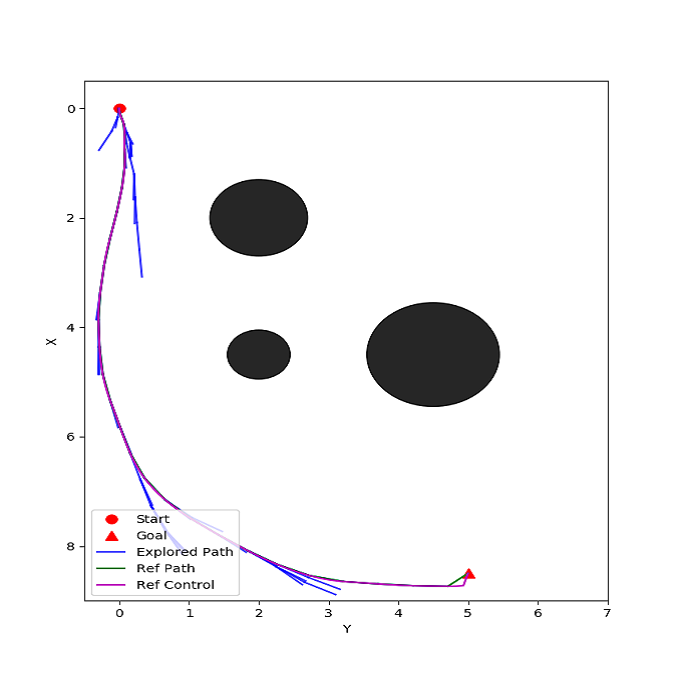}\\
        \mbox{(c) \small$\Delta_1^{max} = 0.3, \Delta_2^{max} = 30$}&\mbox{(d) \small$\Delta_1^{max} = 0.5, \Delta_2^{max} = 50$}\\
    \end{array}$
    \caption{Comparative analysis: [a - d] showcase the performance of Robust RRT-KBF, with increasing uncertainties.}
    \label{fig:inc_unc}
\end{figure}

We can also control the degree of uncertainty in our simulation by setting appropriate values for $\Delta_{1}^{max}$ and $\Delta_{2}^{max}$ (\ref{eq:cbf_uncertain_max}). The designed Robust RRT-KBF controller behaves as RRT-KBF controller on setting uncertainty factor $\Delta_{1}^{max} = \Delta_{2}^{max} = 0$. On increasing the uncertainty i.e. $\Delta_{1}^{max}$ and $\Delta_{2}^{max}$, we can observe that the path planner strictly avoids the obstacle cluster, see Fig.~\ref{fig:inc_unc}.

\subsection{Performance Analysis}

One of the primary focuses of this work is to ensure both path safety and stability without introducing significant computation overhead on the existing RRT-based path-planner. Table \ref{tab:comp} shows the contrast between the RRT, RRT-CBF QP, and the proposed RRT-KBF algorithm with an average run time of over 100 simulations in four different scenarios. While the RRT algorithm is the fastest, it only enforces safety constraints. Using the QP-solver to iteratively find the optimal path for dynamic robots proves to be a costlier approach in terms of time complexity. The proposed RRT-KBF algorithm is able to introduce safety, kinematic, and dynamic constraints on RRT with minimal computational overhead. In comparison with \cite{CBF_RRT}, the use of random sampling in place of QP-solver preserves the random nature of RRT and makes the algorithm computationally efficient.

When we introduce model uncertainty into the planning framework, the performance of the algorithm is directly dependent on the set of uncertainty bounds, as seen in Table \ref{tab:rob_comp}. By setting $\Delta_1^{max} = \Delta_2^{max} = 0$, the Robust RRT-KBF effectively becomes RRT-KBF and gives the best run-time. On increasing the uncertainty parameters $\Delta_1^{max}, \Delta_2^{max}$, the algorithm run-time increases significantly. Although the run-time of Robust RRT-KBF is significantly higher under high uncertainty, this is faster when compared to the traditional RRT-CBF QP.

\section{CONCLUSIONS} 
\label{S5}
In this paper, we have presented a new technique for a safety-critical path planner. The novelty of our solution unlike the conventional alternatives is usage of Rapidly Exploring Random Trees (RRT) and Kinodynamic Barrier Functions (KBF) for efficient planning. We also showcased robustness of RRT-KBF in motion planning, and used the systemic nature and flexibility offered by them to address uncertainties within the model. Further, we also illustrated that by combining robust motion planning along with real-time safety-critical control not only improves high-level planning but even real-time collision avoidance for dynamic robotics. 
\bibliographystyle{rrt_kbf}


\balance
\bibliography{rrt_kbf}



\end{document}